# Benchmarking Adaptive Intelligence and Computer Vision on Human-Robot Collaboration

Salaar Saraj[1], Gregory Shklovski[1], Kristopher Irizarry[1], Jonathan Vet[1], Yutian Ren[1]
[1]California Institute for Telecommunications and Information Technology,
University of California, Irvine

*Abstract*- **Human-Robot Collaboration (HRC) is vital in Industry 4.0, using sensors, digital twins, collaborative robots (cobots), and intention-recognition models to have efficient manufacturing processes. However, Concept Drift is a significant challenge, where robots struggle to adapt to new environments. We address concept drift by integrating Adaptive Intelligence and self-labeling (SLB) to improve the resilience of intention-recognition in an HRC system. Our methodology begins with data collection using cameras and weight sensors, which is followed by annotation of intentions and state changes. Then we train various deep learning models with different preprocessing techniques for recognizing and predicting the intentions. Additionally, we developed a custom state detection algorithm for enhancing the accuracy of SLB, offering precise state-change definitions and timestamps to label intentions. Our results show that the MViT2 model with skeletal posture preprocessing achieves an accuracy of 83% on our data environment, compared to the 79% accuracy of MViT2 without skeleton posture extraction. Additionally, our SLB mechanism achieves a labeling accuracy of 91%, reducing a significant amount of time that would've been spent on manual annotation. Lastly, we observe swift scaling of model performance that combats concept drift by fine tuning on different increments of self-labeled data in a shifted domain that has key differences from the original training environment.. This study demonstrates the potential for rapid deployment of intelligent cobots in manufacturing through the steps shown in our methodology, paving a way for more adaptive and efficient HRC systems.**

## I. INTRODUCTION

The field of Human Robot Collaboration (HRC) aims to bring together humans and robots toward a unified goal. Industry 4.0 has driven the adoption of sensors, digital twin modeling, collaborative robots, and machine learning. Sensor systems have enabled manufacturers in automotive, aerospace, and agriculture industries to automate processes and make data driven enhancements [1]. Digital Twin simulations provide manufacturers opportunities to rapidly simulate and decrease throughput in human-centered assembly tasks [2]. Collaborative robots (or cobots) are often configured to recognize human actions or intentions. Intention-recognition models are machine learning classifiers that are trained on videos of assembly tasks. A pressing hurdle in the adoption of HRC is the lack of robustness when adapting to new tasks and environments, known as Concept Drift. Adaptive Intelligence, which specializes in learning or performing with out-of-distribution datasets, is an active area of research with multiple approaches.

Existing strategies to address concept drift include 1. increasing the robustness of training data [3], and 2. modifying the architecture of robot intelligence models to improve resilience to out-of-distribution (OOD) data [4].

Generating large datasets that encompass a plethora of tasks and environments assists robots with generalizing to OOD tasks [5]. Another modification to training data that improves robustness against shifting environments is placing the robot's "eye" or camera on the robot's arm [6]. This method is able to disregard context shifts by creating data that is context-agnostic.

However, adjustments in intention-recognition model architecture have also resulted in context-agnostic models that can ignore noise introduced by changes in RGB or depth camera inputs. For instance, 3D skeleton based pose-estimation models provide robustness to changes in camera angle, and lighting of an assembly setting [7]. Similarly, digital twin representations of robots have provided similar strength in denoising input data upon adapting to new contexts [8].

Self Labeling (SLB) models are a novel Adaptive Machine Learning (ML) approach to dataset creation which adapts to context drift in HRC systems by using feedback from temporal causality in the system to create labeled data [9]. This paper proposes the adoption of SLB algorithms for fine-tuning intention-recognition models in HRC assembly tasks. We demonstrate SLB label generation after environment changes in an assembly task. Then we evaluate training and performance across several action recognition models. Upon fine-tuning, we orchestrate the model into our HRC system, and perform an assembly with the adapted model predictions

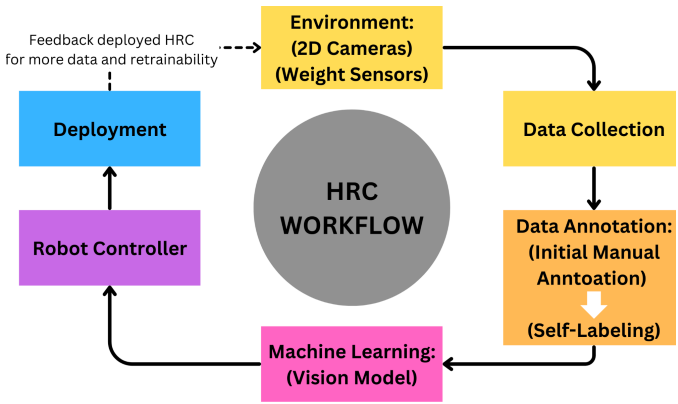

Figure 1: HRC Workflow Diagram

## II. METHODOLOGY

The main task is laid out to quickly incorporate a human robot collaboration (HRC) enabled workspace into any manufacturing or human action driven environment. The steps include data collection strategies and setup, generating self-labeled annotations, and machine learning training techniques and pipelines.

A. Environment:

When it comes to the first initial set up, it is important to define all variables in the environment that are required, that are rooted in the standard operating procedure (SOP) and its requirements. Key parts and tools that are required for the manufacturing process are tracked with weight sensors. However, the most important for this study is a 2D camera, with coverage of all variables that need to be tracked. Our study has discovered that the more careful this initial process is done, the clearer data collection, annotation, and machine learning process becomes. Using devices to track both visual and physical data is vital, and a 2D camera and weight sensors are practically required for this kind of information at the very least. This is because state changes can be accurately detected using weight sensors, while action recognition—particularly for identifying human intentions—requires visual data, as it cannot be reliably achieved without capturing and analyzing actions through visual information.

B. Data Collection:

The next important step is to establish a data collection stream with time stamped information. Accurate timestamps for each data point are essential in this HRC setup, as they ensure precise synchronization when the data is analyzed later. The more precise the synchronization, the more accurate the resulting data labels will be. Data collection devices are connected to a Raspberry Pi and streamed to a centralized source. Given the potential for stream delays or lost frames, maintaining timestamps for every data retrieval is critical to preserving the integrity of the dataset.

C. Data Annotation:

    **a. Manual Annotation:** After understanding the raw data, state changes (e.g., large changes in weight sensors) and their purposes are defined. Then, for each state change there are intentions—the causes or indications for upcoming state changes. These definitions are precise and consistent, making sure there is a meaningful cause and effect relationship. Annotation is made easier by using multi-modal tools synchronized by timestamps, such as charts or videos, with features like pause, rewind, and fast forward. All intentions have the same length to provide a consistent input structure for the model. Post-processing metrics (e.g., time between state changes or intentions) help validate definitions and identify any necessary adjustments. To reduce time consuming manual data annotation, strategies like self-labeling can reduce the overall manual workload. Fully annotated samples are created to train a self-labeling algorithm [9].

    **b. Generating SLB Annotations:** The SLB process requires a state detection algorithm using weight sensor data [9]. The algorithm extracts timestamps of state-changes based on features from the weight sensor, and filters out noise or slight human errors from data collections. We then train the SLB algorithm's interaction time model (ITM) on manually labeled data. Once trained, the SLB can efficiently label the rest of the data, reducing the need for manual labeling [9].

D. Machine Learning:

For machine learning, using a vision model with pretrained weights is very important moving forward, as a large purpose of this task comes from a quick deployment of intelligence in HRC, so using a model that has already been trained on lots of action recognition data is efficient. Models pretrained on the Kinetics dataset [10] work well for this kind of task. Kinetics is a large-scale dataset on 306,245 samples of action recognition. With those weights, using a Multiscale Vision Transformer V2 (MVit2) for action recognition demonstrates a lot of strengths [11]. Transformer models have become a cornerstone in AI, significantly improving tasks like natural language processing, machine translation, and image recognition. Their attention-based learning on sequential information fits well with the idea of SOP in manufacturing, and understanding intentions.

The model is designed to predict and detect intentions from video clips. Annotated data includes numerous labeled intention samples for training. The model also predicts non-intentions (negative samples), which are extracted from video segments outside the intention timestamps. The extraction and quantity of negative samples can be a discussion point, as non-intentions are more complex due to their variance. In this study, we used the same number of negative samples as each intention for our training purposes..

E. Robot Controller:

We forward recognized intentions to our Robot Controller that reads intention labels and dispatches a robot action associated with the given intention label. It follows a client-server architecture where the Robot Controller script is a client that sends instructions to the robot. We define each robot action as an ordered series of arm and gripper movements. The constraints of a well defined path prevent the robot's built-in inverse kinematics from choosing a wrong movement path when given a waypoint that has multiple ways of being reached. The intermediate steps consist of a variable number of approach positions, a grip position, a variable number of post-pickup positions, a handoff position, and a reset position. To generate, send, and process robot execution paths, we created 3 modules (1. Waypoint Calibration, 2. Client, 3. Server.) Before calibrating waypoints, we define the number of intermediate steps that each execution path contains.

The waypoint calibration module consists of a script that reads real-time robot data into a csv file each time a *record* button is pressed. A user will set a robot in freedrive mode, and run the calibration script for each step of the robot's path while moving the robot and pressing *record* at each waypoint.

1. Waypoint Calibration:

Our system depends on csv files containing the waypoints of all execution paths. To generate these csv files, we wrote a script that monitors the robot's position in freedrive mode, and writes the coordinates of a position into a csv, each time a position is recorded. Such a script simplified our calibration process to the following short steps.

1. Start script with "waypoint name" argument
2. Move to desired position in freedrive mode
3. *Record* the current position into a CSV.
4. Repeat these steps for each associated point

2. Robot Controller - Client:

The client is responsible for establishing a connection to the Robot, monitoring execution readiness and completion.

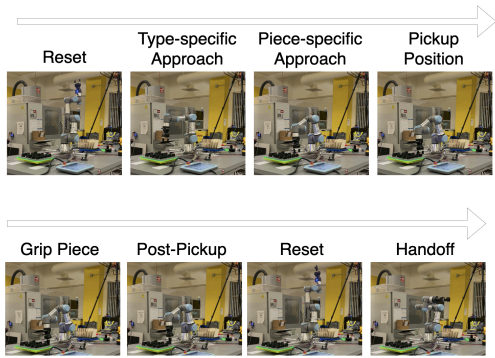

Figure 2: Standardized Sequence of Setpoints

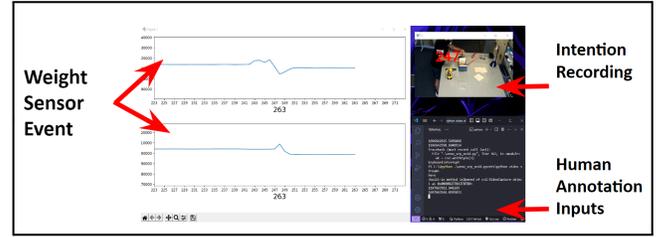

Figure 3: Manual Data Annotation

Once we recorded all of the waypoints into csv files, our robot controller parsed them into lists of set-points that are indexable by Intention Label. When an action method is called on the robot controller the Robot Controller constructs an ordered list of setpoints and then calls a function to iteratively send them to the robot.

3. The Robot Control Loop - Server:

The robot is responsible for receiving set-point arrays, moving to those points, and notifying the client once the movement is complete. We include the grip command as a boolean value that is sent along with the set-point array. The robot receives commands by running a loop that reads new positions from the client and writes them to shared memory on the robot. A loop in a second thread then dispatches movements to those positions, and reports the completion of movements.

Once all the parts explained are assumed to be completed, our study shows that a full system can be created to facilitate the HRC work structure that is being built up. Real time video data is sent to a server or device that stores the model with its trained weights. Then the model will keep making predictions on the video samples, and if a certain intention is detected, then the particular state change that follows that intention is instructed to the robot. Additionally, auxiliary sensor information (like change in joint currents) can be used as a safeguard or progress tracker in the case that the model might make incorrect predictions. We continue collecting data even in the fully deployed system for annotation and re-training purposes in the future.

III. EXPERIMENT

Our study applies every step outlined in the methodology. For our assembly task, we chose building a small wooden chair, which involves connecting wooden dowels, plates, and plastic joints with a screwdriver. The goal is to detect user intentions during the build, enabling the robot to prepare the required pieces, eliminating the need for the user to retrieve them from the tray.

For data collection, we used a 2D PiCam and two weight sensors, each connected to a Raspberry Pi. The video stream was transmitted using OpenCV with a Gstreamer backend over an STP server connection to the data storage device. Each video frame and weight sensor data index included live timestamps. One weight sensor tracked wooden parts, the other plastic. The user builds the chair while the camera and sensors record the process. We collected 201 samples.

For data annotation, we created a script to sync data from both weight sensors and video based on timestamps. The script included controls for adding or deleting labels, rewinding, and reviewing current labels. We labeled all weight sensor events as state changes and defined a specific intention for each. For instance, the state change of picking up the second dowel and floor joint had the intention of attaching the first chair leg. Intentions were consistently defined as 4-second videos for uniform processing. We defined 13 different intentions and a zero sample non-intention, thus creating 14 different classes of prediction for action recognition. We calculated metrics like state durations and intervals between states and intentions to assess the quality of our annotations. One crucial metric to keep track of is making sure there is enough time before and after your intentions between the state changes, so that there is enough time for the intention to be recognized and responded to. Therefore, we adjusted some intention definitions to occur either earlier or later to have enough padding time between state changes.

Before model training, we created scripts to ensure all metrics were calculated and intentions were correctly extracted from the annotated data. This step is crucial for identifying faulty data collections and annotations, leading to more consistent model training and the exploration of edge cases. Additionally, we developed a script to analyze raw data and extract negative samples or non-intentions outside the labeled timestamped intention intervals.

For model training, we used the MViT2 [11] Multi-Scale Transformer with pre-trained weights from Kinetics400 [10], which aligned well with the chair assembly actions [8]. We experimented with various hyperparameters and layer freezing to optimize accuracy in detecting intentions. Recognizing the relevance of user movements, we also integrated skeleton posture data, extracted using Mediapipe, into the MViT2 classification layer [11]. The skeleton posture information is also useful when it comes to domain shifts for the environment. To benchmark our results, we trained a Slow-Fast model [12] with pre-trained Kinetics weights, also incorporating skeleton posture information into the classification layer [10].

Additionally, to combat domain shift due to various workspace setups for HRC, we also used a self-labeling algorithm [9]. We trained the interaction time model(ITM) on our manual labels and created our own custom state change detection algorithm on the weight sensor collections in order to identify timestamps of every state change. Our state detection algorithm precisely found timestamps of state changes, by processing raw weight data by first normalizing and smoothing it to reduce noise. It then detects significant changes in the smoothed data, corresponding to different physical states or actions, using predefined patterns and thresholds. A lookahead mechanism ensures that detected changes are sustained, confirming valid state transitions. This method enables accurate identification of states in complex weight data. After that, we used self-labeled data on new data collections in a completely different HRC environment and fine tuned our models to the newer data. The environment was in a different room, with a different table, weight tray, robot positioning set up, and camera angle. The fine-tuning process involves using our trained weights from the first dataset, with a lower learning rate than original training to be fit with the small sample size of the new data on 30 epochs.

The robot we chose for our experiment was Universal Robots UR-E3 6 DOF arm, which uses the UR script programming language to command robot movements, read robot state, and receive instructions over a network.

To program our robot, we first created a waypoint calibration script in python with the *RTDE Python Client Library* [13]. Next, we determined a list of intermediate waypoints, and created csv files to store each one.

Generally, the robot uses UR Inverse Kinematics to place a tool at a desired position, which means there are multiple arm poses that can correspond to the same waypoint. To standardize the poses, we enforced a fully extended pose at the start of each full execution path, which eliminated ambiguities for interpolating the execution path. Each csv file contains a list of all assembly part positions in a standardized order.

To direct the robot to help with a specific intention, we first wrote a script to capture robot Tool Center Points (TCP). After recording our desired waypoints into a separate CSV for each waypoint type (ie. pickup, approach, handoff, etc.), we wrote a Robot Controller class that uses the ur_rtde[14] library. We chose this library because of its fine grained control over data being sent to and from the robot such as set-point arrays and watchdog booleans that represent if a command is completed or in-progress.

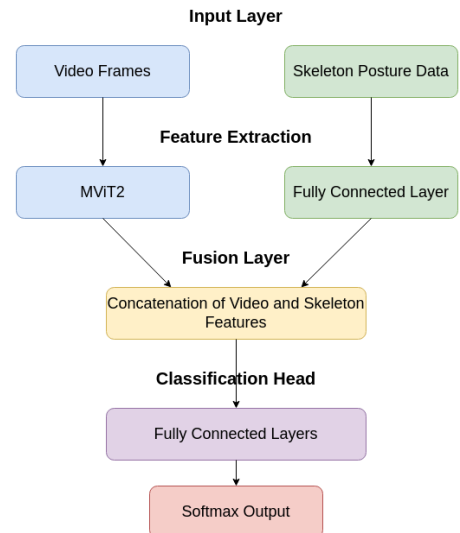

Figure 4: MViT2Skel Model

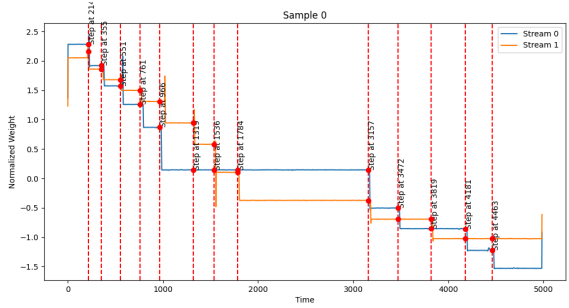

Figure 5: State Change Algorithm Labels

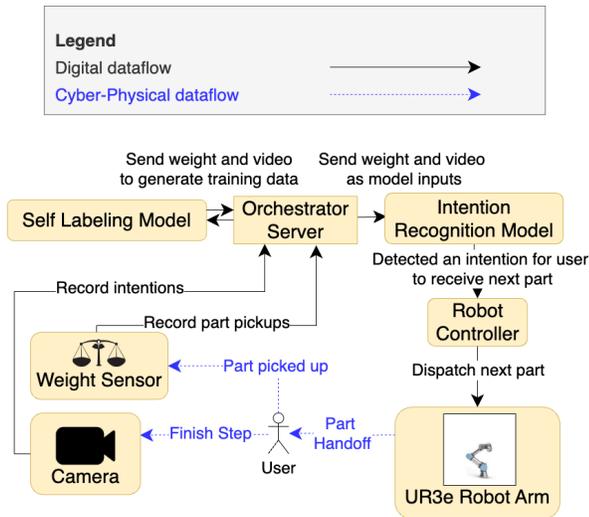

Figure 6: System Architecture

| Model | Accuracy |
|---|---|
| MViT2 base model | 79% |
| Slow-Fast model with skeleton posture preprocessing | 69% |
| MViT2 model with skeleton posture preprocessing | 83% |

Figure 7: Action Recognition Performance

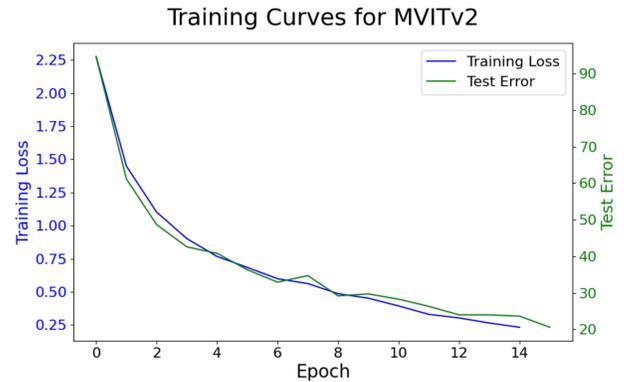

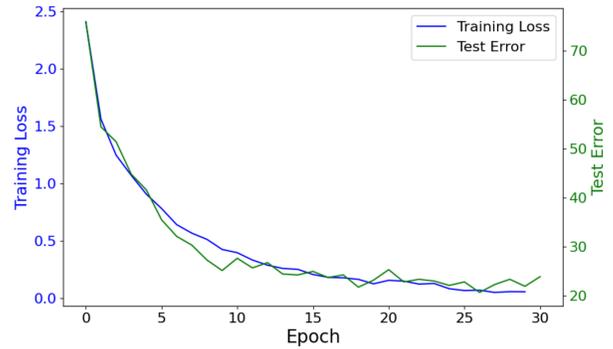

Figure 8: Training Curves Prior to Concept Drift

Our multithreaded implementation of the robot URP used one thread to read setpoint arrays whenever the client toggled a flag. Along with the set point array, the robot also reads a boolean that represents desired gripper state, 1 for close and 0 for open. When a new move is read, it is copied into a setpoint variable that is shared with the second thread. Upon reading a new setpoint, the second thread continuously directs the robot to the given tool center point using a MoveJ command, and then toggles the watchdog flag to signal to the client that a move is completed.

To facilitate that full demonstration of all the parts of our study, we set up a server to host the trained model and use a client-server connection to the 2D camera monitoring the HRC workspace. We use a continuous input of 4 second videos into the trained model, and based on the model predictions we then give out the corresponding set of instructions to the collaborative robot. We create a 2 second crossover for the videos we send to the model to make sure we capture everything that is required to get the right predictions on our intentions. Our weight sensors act as a secondary check on the current progress of the build, in terms of the state-changes and what pieces have been removed. If the model we are testing requires the skeleton posture extraction, we do so at the server side before sending it into the model.

## IV. RESULTS

Our model performance after our training offered the following promising results for the initial training lab space.

The MViT2Skel model achieved an accuracy of **83%**, outperforming MViT2 by 4%, on the 14 different action recognition classes throughout this specific HRC process. After adjusting the self-labeling mechanism with a custom weight sensor state detection algorithm, we improved the self-labeling accuracy to **91%** [9] . Lastly, to test the capabilities of our HRC ecosystem with a large domain shift, we collected 20 new data collections in a completely different robot lab, with various set up differences. Initially, we fine-tuned our MViT2Skel model on the first 10 collections, and got an average cross-validation accuracy of **30%** over 40 epochs. Then, we fine-tuned it including all 20 samples, with an average cross-validation **45.2%**.

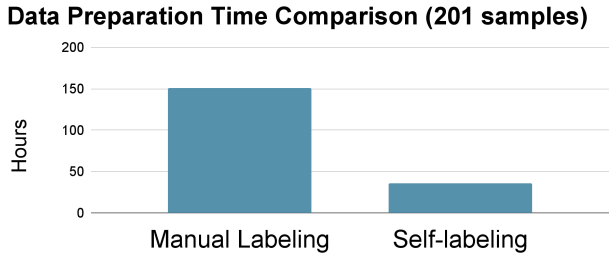

Figure 9: Total data preparation time consumption comparison over 201 samples. This includes time taken up during data collection and time labeling data.

## V. DISCUSSION

The various outcomes of our experiments show that the use of adaptive intelligence paired with deep learning will eliminate the time and labor required to generate annotated datasets when adapting robots to new contexts. We hope that reducing perceived cost of fine-tuning robot intelligence models while maintaining fine-tuned performance will encourage quicker development of HRC manufacturing systems and greater adoption of Industry 4.0 automation techniques.

Improving MViT2's [11] base model trained on our data by using skeleton posture preprocessing showed that changing the model architecture to classify using additional features is very useful and that MViT2Skel model architecture can be used more in the future. Additionally, achieving an 83% accuracy on 14 different classes with only 201 data samples that we collected shows that every manufacturing space has the ability to quickly engage themselves with predictive intelligence for robots in their workspace without extreme numbers of data. This is leveraged by using pre-trained weights from Kinetics400 and feature extraction of skeleton posture information.

Additionally, another motivating factor to incorporate predictive intelligence in HRC comes from the efficiency in data labeling in this study. Using a self-labeling strategy [9] on our data, and creating a specific state detection algorithm that captures state changes to near perfect accuracy, showed a very powerful and efficient way to create more labeled data without any manual annotation [7]. The 91% similarity to our manual annotations shows that an extreme amount of time can be saved if this labeling strategy is used more in the future in various HRC or general action recognition fine-tuning environments. In our case, using self-labeling would have saved around 30 minutes/label which translates to saving 100 hours spent on data annotations for the 201 samples [9].

Furthermore, using that same self-labeled strategy on a small sample size of data in a completely new environment offers a view of how quickly action recognition can battle domain shifts. Seeing a 15% accuracy increase when comparing 10 to 20 samples of self-labeled data in a new environment, shows that the model can learn quickly in the new environment while eliminating the high amount of labor required when traditionally creating a dataset of an appropriately large sample size. Therefore, similar action recognition tasks that may be performed in different labs can be fine-tuned very quickly by using the MViT2Skel model and self-labeling [9]. This is driven by the flexibility of having a more constant feature during a domain shift by using skeletal information, while also creating fully labeled data in the matter of seconds after collection using the self-labeling strategy [9].

We were limited by certain factors in this study, largely coming from using a mixed sequence of intentions, lack of high frequency robot control, and the complete accuracy of action recognition is still not ready for full dependence on robots, requiring some precautionary procedures that can handle cases where there are missed predictions. In the future with this work, we hope to create an environment where multiple human assemblies can share the same robot, so we can expand off the limitation of fixed sequences. We also wish to explore different HRC environments that can use a variety of different orders in their steps to also evaluate performance on non-fixed sequences. Also, we would do even more different domain shifts with more self-labeled data to try to see how much we can leverage the data-efficiency of our system through fine-tuning. Additionally, we wish to still decrease as much labeled sample required to reach high accuracies through different adaptive intelligence techniques, which could include even more forms of feature extraction.

## VI. CONCLUSION

All these improvements demonstrate significant advances in action recognition for human-robot collaboration (HRC), particularly in terms of data collection, annotation efficiency, and the ability to handle domain shifts through fine-tuning. Achieving high accuracy with just 201 samples underscores the feasibility of quickly deploying this system across various manufacturing environments. By following the steps outlined in our methodology, organizations can enhance the performance of action-recognition-driven HRC systems, substantially reducing both time and effort in generating annotated datasets. The results indicate that such systems with high data-efficiency can significantly reduce operational costs and increase adaptability, which are critical to the adoption of Industry 4.0 techniques.

Future research can explore further optimizations, including expanding to multi-robot environments, handling more complex domain shifts, and investigating additional forms of adaptive intelligence to further decrease the need for labeled data and improve generalization capabilities.

## VII. ACKNOWLEDGEMENTS